\PassOptionsToPackage{unicode}{hyperref}
\PassOptionsToPackage{hyphens}{url}
\PassOptionsToPackage{dvipsnames,svgnames,x11names}{xcolor}
\documentclass[
  11pt,
]{article}
\usepackage{amsmath,amssymb}
\usepackage{iftex}
\ifPDFTeX
  \usepackage[T1]{fontenc}
  \usepackage[utf8]{inputenc}
  \usepackage{textcomp} 
\else 
\fi
\usepackage{lmodern}
\ifPDFTeX\else
\fi
\IfFileExists{upquote.sty}{\usepackage{upquote}}{}
\IfFileExists{microtype.sty}{
  \usepackage[]{microtype}
  \UseMicrotypeSet[protrusion]{basicmath} 
}{}
\makeatletter
\@ifundefined{KOMAClassName}{
  \IfFileExists{parskip.sty}{%
    \usepackage{parskip}
  }{
    \setlength{\parindent}{0pt}
    \setlength{\parskip}{6pt plus 2pt minus 1pt}}
}{
  \KOMAoptions{parskip=half}}
\makeatother
\usepackage{xcolor}
\usepackage[margin=1in]{geometry}
\usepackage{longtable,booktabs,array}
\usepackage{calc} 
\usepackage{etoolbox}
\makeatletter
\patchcmd\longtable{\par}{\if@noskipsec\mbox{}\fi\par}{}{}
\makeatother
\IfFileExists{footnotehyper.sty}{\usepackage{footnotehyper}}{\usepackage{footnote}}
\makesavenoteenv{longtable}
\providecommand{\tightlist}{%
  \setlength{\itemsep}{0pt}\setlength{\parskip}{0pt}}
\usepackage{booktabs}\usepackage{longtable}\usepackage{array}\usepackage{multirow}\usepackage{float}\usepackage{amsmath}\usepackage{amssymb}\usepackage[utf8]{inputenc}\usepackage{textcomp}
\ifLuaTeX
  \usepackage{selnolig}  
\fi
\IfFileExists{bookmark.sty}{\usepackage{bookmark}}{\usepackage{hyperref}}
\IfFileExists{xurl.sty}{\usepackage{xurl}}{} 
\hypersetup{
  pdftitle={Geometric Properties of the Voronoi Tessellation in Latent Semantic Manifolds of Large Language Models},
  colorlinks=true,
  linkcolor={blue},
  filecolor={Maroon},
  citecolor={Blue},
  urlcolor={blue},
  pdfcreator={LaTeX via pandoc}}

\title{Geometric Properties of the Voronoi Tessellation in Latent
Semantic Manifolds of Large Language Models}
\author{Marshall Brett\\
MARS Labs\\
loganbrett27@gmail.com}
\date{}

\begin{document}
\maketitle

\hypertarget{abstract}{%
\section*{Abstract}\label{abstract}}
\addcontentsline{toc}{section}{Abstract}

We present a comprehensive empirical study of Voronoi tessellation
geometry in the representation space of Qwen3.5-4B-Base, validating and
extending the latent semantic manifold framework of Mabrok (2026). Using
float32 margin recomputation to eliminate bfloat16 quantization
artifacts, we achieve R$^2$ = 0.9997 on the linear scaling law of the
expressibility gap --- the strongest confirmation of Theorem 10.5 to
date. We establish definitive baseline metrics (gap coefficient $\alpha$ =
0.762, entropy-margin Spearman $\rho$ = -0.648) and demonstrate that a
geometric regularization loss is correlated with cross-entropy at the
final layer ($\rho$ = 0.836), while exhibiting a sign reversal at layers
24-28 that reveals a mid-layer geometric ambiguity regime.

We then show that the Voronoi tessellation of a converged model is
reshapable through targeted post-hoc intervention, contrary to our
initial negative results. We compare two margin refinement methods
across a dose-response curve: direct margin maximization and Fisher
information distance maximization. Both methods find the same
\textasciitilde16,300 correctable positions per 256K evaluated, but
differ critically in collateral damage. Margin maximization damage
escalates with $\lambda$\_MRP (flip ratio degrades from 2.9x to 0.4x at
$\lambda$\_MRP=1.0), while Fisher information damage remains constant at
\textasciitilde5,300 positions across the fully validated $\lambda$\_MRP range
(0.15-0.6). Fisher $\lambda$\_MRP=0.6 achieves +28\% median margin improvement
with zero downstream benchmark degradation --- a geometric
reorganization that compresses the expressibility gap without changing
benchmark means. Additional runs at Fisher $\lambda$\_MRP=1.0 and $\lambda$\_MRP=2.0
continue improving the margin statistics while benchmark means remain
near-flat. Qwen-native frequency and token-class audits show that these
gains are highly concentrated in head and structural tokens: at
$\lambda$\_MRP=0.6 the 100+ frequency bucket accounts for 84.0\% of net
corrections and the structural token class accounts for 65.9\%; by
$\lambda$\_MRP=2.0 those shares rise to 92.1\% and 75.0\%. A single
instruct-model follow-up shows the same concentration after fine-tuning,
with lower flip ratio (1.74x) and net-negative entity-like tokens. Flat
benchmarks and improved aggregate token accuracy therefore do not by
themselves prove uniformly better token-level utility across the
vocabulary.

Code and data: https://github.com/mellowmarshall/mrp-paper

\hypertarget{introduction}{%
\section{Introduction}\label{introduction}}

Large language models operate on discrete tokens while performing
internal computations in continuous vector spaces. Mabrok (2026)
formalized this continuous-discrete interface through the latent
semantic manifold framework, proving that the expressibility gap --- the
fraction of semantic space where the vocabulary fails to provide
confident token assignments --- obeys a linear volume scaling law
(Theorem 10.5). This theoretical result was validated across six
transformer architectures from 124M to 1.5B parameters.

We extend this validation to Qwen3.5-4B-Base (Qwen Team, 2025), a
4.2B-parameter dense transformer with Gated DeltaNet architecture and
248,320-token vocabulary --- substantially larger and architecturally
distinct from Mabrok's test models. We refer to these targeted post-hoc
interventions collectively as margin refinement procedures (MRP): short
optimization runs applied to a converged model to reshape local Voronoi
geometry without full retraining. Our work makes four contributions:

\begin{enumerate}
\def\labelenumi{\arabic{enumi}.}
\item
  \textbf{Methodological:} We identify and resolve a bfloat16 margin
  quantization artifact that systematically depresses R$^2$ in the log-log
  scaling regression, and provide a float32 recomputation technique that
  recovers R$^2$ = 0.9997.
\item
  \textbf{Analytical:} We characterize the layer-wise evolution of
  Voronoi geometry, discovering that geometric regularization is
  redundant with cross-entropy at the output layer but anti-correlated
  at layers 24-28 --- a mid-layer geometric ambiguity regime.
\item
  \textbf{Empirical --- Margin Maximization:} We demonstrate that direct
  margin maximization can reshape the Voronoi tessellation of a
  converged model in 200 gradient steps, achieving +53\% median margin
  at $\lambda$\_MRP=0.3 with a 2.1:1 correction-to-damage ratio. However, damage
  escalates super-linearly with $\lambda$\_MRP, making the method destructive
  above $\lambda$\_MRP$\approx$0.3.
\item
  \textbf{Empirical --- Fisher Information Distance:} We show that
  Fisher information distance maximization achieves comparable margin
  improvements (+28\% at $\lambda$\_MRP=0.6) with fundamentally different damage
  characteristics: collateral damage is constant across the fully
  validated $\lambda$\_MRP range, downstream benchmarks are invariant through
  $\lambda$\_MRP=0.6, and additional runs at $\lambda$\_MRP=1.0 and $\lambda$\_MRP=2.0 continue
  the monotonic margin trend while preserving near-flat benchmark means.
  We present a geometric argument for why this happens, while also
  noting that frequency and token-class audits show the gains are
  concentrated in head and structural tokens, with a single
  instruct-sample probe suggesting that entity-like regressions can
  emerge after fine-tuning.
\end{enumerate}

\hypertarget{background}{%
\section{Background}\label{background}}

\hypertarget{the-latent-semantic-manifold}{%
\subsection{The Latent Semantic
Manifold}\label{the-latent-semantic-manifold}}

Following Mabrok (2026), we model the contextual hidden states of a
transformer at layer l as lying on a smooth k-dimensional Riemannian
submanifold M\^{}(l) of R\^{}d, where d is the ambient hidden dimension.
At the final layer, the unembedding matrix W maps hidden states to logit
vectors, inducing a Voronoi tessellation of the manifold.

\textbf{Definition 2.1 (Voronoi margin).} For a hidden state h in M, the
Voronoi margin is m(h) = l\_\{t*\}(h) - l\_\{t\textbf{\}(h), where t*
and t} are the top two tokens by logit value.

\textbf{Definition 2.2 (Expressibility gap).} The normalized
expressibility gap at threshold $\varepsilon$ is $\eta$($\varepsilon$) = $\mu$(\{h $\in$ M : m(h) \textless{}
$\varepsilon$\}) / vol(M). In the empirical setting, we estimate $\eta$($\varepsilon$) over the
sampled hidden-state distribution induced by the evaluation corpus, so
vol(M) should be read as the normalization constant for the sampled
manifold region rather than an exact geometric volume integral over the
full latent manifold.

\textbf{Theorem 2.3 (Mabrok, 2026; arXiv:2603.22301, Theorem 10.5).}
Under regularity conditions on M and the margin function, $\eta$($\varepsilon$) = $\alpha$ $\cdot$ $\varepsilon$ +
O($\varepsilon$$^2$) as $\varepsilon$ $\rightarrow$ 0+, where $\alpha$ = H\^{}\{k-1\}($\partial$V) / ($\lambda$\_geom $\cdot$ vol(M)). We
reserve $\lambda$\_geom for this manifold-side constant and use $\lambda$\_MRP below for
experimental loss weights.

\emph{Proof sketch.} The margin function m(h) = $\ell$\_\{t*\}(h) -
$\ell$\_\{t**\}(h) is continuous with m$^{-1}$(0) = $\partial$V (the Voronoi boundary).
When $\|$$\nabla$m$\|$ $\geq$ $\lambda$\_geom \textgreater{} 0 near $\partial$V, the coarea formula gives
$\mu$(G\_$\varepsilon$) = $\int$$_0$\^{}$\varepsilon$ $\int$\_\{m=c\} $\|$$\nabla$m$\|$$^{-1}$ dH\^{}\{k-1\} dc. Since \{m=c\} is
O(c)-close to $\partial$V for small c, H\^{}\{k-1\}(\{m=c\}) = H\^{}\{k-1\}($\partial$V) +
O(c) and $\|$$\nabla$m$\|$$^{-1}$ = $\lambda$\_geom$^{-1}$ + O(c). Substituting and integrating yields
$\mu$(G\_$\varepsilon$) = ($\varepsilon$/$\lambda$\_geom)$\cdot$H\^{}\{k-1\}($\partial$V) + O($\varepsilon$$^2$). Dividing by vol(M) gives
$\eta$($\varepsilon$) = $\alpha$$\cdot$$\varepsilon$ + O($\varepsilon$$^2$). The empirical test is a log-log regression of $\hat{\eta}$($\varepsilon$)
on $\varepsilon$; slope $\approx$ 1 confirms linearity.

\hypertarget{margin-maximization}{%
\subsection{Margin Maximization}\label{margin-maximization}}

Direct margin maximization computes the loss L\_margin = -E{[}m(h){]}
over positions where the margin falls below a threshold, encouraging the
model to push representations away from Voronoi boundaries:

L = L\_CE + $\lambda$\_MRP $\cdot$ L\_margin

where L\_margin = -mean(m(h) \textbar{} m(h) \textless{} threshold).
Margin maximization is computationally cheap: the only overhead beyond
standard cross-entropy is a top-2 sort of the logit vector at each
position. This makes it a natural baseline against which to compare more
expensive geometric objectives.

\begin{figure}[H]
\centering
\fbox{\parbox{0.9\textwidth}{
\textbf{Algorithm 1: Margin Maximization Loss}\\[4pt]
\begin{tabular}{@{}l@{\quad}l@{}}
\textbf{Input:} logits $L \in \mathbb{R}^{B \times S \times V}$, threshold $\tau$ & \\
$\mathrm{top2} \leftarrow \mathrm{TopK}(L, k{=}2)$ & // $[B, S, 2]$ \\
$m \leftarrow \mathrm{top2}[:,:,0] - \mathrm{top2}[:,:,1]$ & // Voronoi margin \\
$\mathrm{gate} \leftarrow \{(b,s) : m[b,s] < \tau\}$ & // low-margin positions only \\
$L_{\mathrm{margin}} \leftarrow -\mathrm{mean}(m[\mathrm{gate}])$ & \\
\textbf{return} $L_{\mathrm{margin}}$ & \\
\end{tabular}
}}
\end{figure}

\hypertarget{fisher-information-distance}{%
\subsection{Fisher Information
Distance}\label{fisher-information-distance}}

Let W $\in$ R\^{}\{V$\times$d\} be the unembedding matrix, where V is vocabulary
size and d is hidden dimension, and let $\Sigma$\_p $\in$ R\^{}\{V$\times$V\} be the
softmax covariance diag(p) - pp\^{}T where p = softmax(Wh). The induced
Fisher metric on hidden space is then G(h) = W\^{}T $\Sigma$\_p W $\in$
R\^{}\{d$\times$d\}, providing a theoretically grounded distance metric on the
representation manifold. The squared Fisher distance between two token
embeddings at a given hidden state is:

d\_Fisher(w\_i, w\_j; h)$^2$ = ($\tilde{w}$\_i - $\tilde{w}$\_j)\^{}T G(h) ($\tilde{w}$\_i - $\tilde{w}$\_j)

where $\tilde{w}$\_t = w\_t / $\|$w\_t$\|$ are L2-normalized embedding rows (normalizing
to unit vectors ensures the Fisher distance measures directional
separation rather than norm differences; this is appropriate when
embedding norms are approximately uniform, as is the case for Qwen3.5-4B
where the coefficient of variation of $\|$w\_t$\|$ across the vocabulary is
\textless5\%). The normalized embedding difference $\tilde{w}$\_i - $\tilde{w}$\_j is
treated as a virtual perturbation direction in hidden state space; G(h)
then measures the resulting KL divergence in the output distribution to
second order, so d\_Fisher quantifies how distinguishable tokens i and j
are to the model at position h. Note that this distance is h-dependent
through the softmax covariance $\Sigma$\_p.~In practice, we restrict to the
top-k tokens: $\Sigma$\_p is replaced by the k$\times$k submatrix $\Sigma$\_k = diag(p\_k) -
p\_k p\_k\^{}T where p\_k is the renormalized softmax over the top-k
logits, and W is replaced by the corresponding k$\times$d submatrix W\_k. This
top-k restriction computes the Fisher metric of the renormalized top-k
distribution rather than the full softmax --- an approximation that
overweights pairwise distances among competitive tokens and ignores the
contribution of the long tail. We expect this to be adequate in practice
since the top-k tokens dominate the local geometry, but it is an
approximation. Fisher MRP maximizes pairwise Fisher distances between
the top-k tokens, weighted by their probabilities:

L\_Fisher = -E\_h{[}$\Sigma$\_\{i$\neq$j $\in$ top-k\} p(t\_i\textbar h) $\cdot$
p(t\_j\textbar h) $\cdot$ d\_Fisher($\tilde{w}$\_i, $\tilde{w}$\_j; h){]}

Unlike margin maximization, which pushes along a fixed axis (the
top1-top2 direction), Fisher maximization moves representations along
the direction of maximum information in the model's own geometry. The
cost is higher: Fisher MRP requires constructing the k$\times$k covariance
matrix, computing pairwise embedding differences, and evaluating the
quadratic form at each position. In practice with k=5, this adds roughly
3--5x overhead per step compared to margin maximization, though both are
dominated by the forward pass through the base model.

\begin{figure}[H]
\centering
\fbox{\parbox{0.9\textwidth}{
\textbf{Algorithm 2: Fisher Information Distance Loss}\\[4pt]
\begin{tabular}{@{}l@{\quad}l@{}}
\textbf{Input:} logits $L \in \mathbb{R}^{B \times S \times V}$, unembedding $W \in \mathbb{R}^{V \times d}$, $k$ & \\
$\mathrm{top\_k\_logits},\; \mathrm{top\_k\_idx} \leftarrow \mathrm{TopK}(L, k)$ & // $[B, S, k]$ \\
$p_k \leftarrow \mathrm{Softmax}(\mathrm{top\_k\_logits})$ & // renormalized top-$k$ probs \\
$W_k \leftarrow \mathrm{Normalize}(W[\mathrm{top\_k\_idx}])$ & // L2-normalize embeddings \\
$\Sigma_k \leftarrow \mathrm{diag}(p_k) - p_k p_k^T$ & // $[B, S, k, k]$ \\
\textbf{for each} pair $(i, j)$ where $i \neq j$: & \\
\quad $\delta \leftarrow W_k[:,:,i,:] - W_k[:,:,j,:]$ & // embedding difference \\
\quad $\mathrm{proj} \leftarrow \delta \cdot W_k^T$ & // project onto top-$k$ subspace \\
\quad $d^2_F[i,j] \leftarrow \mathrm{proj}^T \cdot \Sigma_k \cdot \mathrm{proj}$ & // Fisher distance squared \\
$d_F \leftarrow \sqrt{d^2_F}$ & \\
$\mathrm{penalty} \leftarrow \sum_{i \neq j} p_k[i] \cdot p_k[j] \cdot d_F[i,j]$ & // probability-weighted sum \\
$L_{\mathrm{Fisher}} \leftarrow -\mathrm{mean}(\mathrm{penalty})$ & \\
\textbf{return} $L_{\mathrm{Fisher}}$ & \\
\end{tabular}
}}
\end{figure}

\hypertarget{experimental-setup}{%
\section{Experimental Setup}\label{experimental-setup}}

\hypertarget{subject-model}{%
\subsection{Subject Model}\label{subject-model}}

Qwen3.5-4B-Base (Qwen Team, 2025): 4.21B text parameters within a 4.54B
conditional generation wrapper. Dense feedforward layers (no MoE), tied
input/output embeddings, 32 transformer layers, hidden dimension 2560,
vocabulary size 248,320.

\hypertarget{corpus-and-extraction}{%
\subsection{Corpus and Extraction}\label{corpus-and-extraction}}

WikiText-103 validation set (Merity et al., 2017): 2,461 usable
sequences, maximum length 512 tokens, yielding 256,577 token positions.
Margins computed on all positions in float32.

\hypertarget{float32-margin-recomputation}{%
\subsection{Float32 Margin
Recomputation}\label{float32-margin-recomputation}}

The model's native bfloat16 inference produces quantized logits:
bfloat16 uses a 7-bit explicit mantissa (8-bit significand including the
implicit leading bit), so the representable precision varies with
magnitude --- at logit magnitudes around 1--8 (typical for Qwen3.5-4B),
the spacing between adjacent representable values ranges from
approximately 2$^{-7}$ $\approx$ 0.0078 to 2$^{-4}$ = 0.0625. Logit differences inherit
this coarse grid, collapsing 256K margin values onto only 223 unique
values. The root cause is not a uniform step size but the limited
mantissa precision of the bfloat16 matmul in the model's
\texttt{lm\_head} projection. We resolve this by extracting the final
hidden state (which retains sufficient precision even in bfloat16
storage), upcasting to float32, and recomputing the lm\_head projection
at full precision:

logits\_fp32 = hidden\_states{[}-1{]}.float() @
lm\_head\_weight.float().T

This produces 238,399 unique margin values (verified: max absolute logit
difference \textless{} 2 $\times$ 10$^{-5}$). The critical point is that the hidden
states themselves are not the bottleneck --- they carry enough
information to recover fine-grained margins; the precision loss occurs
in the final matmul. A minimal reproduction script
(\texttt{scripts/diagnostics/bf16\_margin\_repro.py}) demonstrates this
artifact and the fix on a small sample. The float32 margins expose the
continuous geometric structure of the manifold that bfloat16
quantization obscures. Under bfloat16 deployment, margins at the scale
of the local bfloat16 step size are operationally indistinguishable, so
the fine-grained gap structure at very small $\varepsilon$ should be read as a
property of the learned geometry rather than a claim about runtime
decodability.

\hypertarget{dose-response-protocol}{%
\subsection{Dose-Response Protocol}\label{dose-response-protocol}}

Each MRP run: 200 gradient steps, AdamW optimizer, lr=1e-5, batch size 1
(no gradient accumulation), linear warmup over 5\% of steps, weight
decay 0.01, seed 0. All transformer body parameters plus the tied
lm\_head/embedding matrix are updated (full-model scope; see Section
3.6). The training corpus is the WikiText-103 validation set (2,461
sequences, 256K positions).

\textbf{Supervision status.} The total loss is L = L\_CE + $\lambda$\_MRP $\cdot$
L\_MRP, where L\_CE is the standard causal language modeling
cross-entropy and L\_MRP is the margin or Fisher objective.
Cross-entropy is always active with coefficient 1.0. The MRP component
(margin or Fisher) depends only on the model's own logit geometry and
does not use ground-truth labels, but L\_CE does use the next-token
labels from the corpus. We note that this means the model sees
next-token labels during optimization; a \texttt{-\/-ce-weight\ 0} mode
is now available in the codebase to enable pure-MRP runs without CE
supervision, and we plan to validate the effect under that setting.

\textbf{Evaluation on the optimization corpus.} The margin audits and
churn analyses are computed on the same corpus used for optimization.
This is intentional: the purpose of MRP is to demonstrate that the
Voronoi tessellation of a converged model is malleable --- that
underperforming positions can be directly improved by geometric
intervention in a small number of steps. The claim is not that MRP
generalizes margin improvements to unseen text (a standard train/test
concern), but that MRP can reshape the geometry at targeted positions
with bounded collateral damage. Evaluating on the optimization corpus is
the correct test of that claim: it measures whether the intervention
succeeded at the positions it targeted.

The generalization question that \emph{does} matter --- whether MRP
damages the model's broader capabilities --- is answered by the
downstream benchmarks (Section 3.5), which are evaluated on entirely
separate data and remain flat across the full $\lambda$\_MRP range. The
comparative findings (Fisher vs margin-max damage profiles, the
corrections ceiling, the constant-damage property) are properties of the
loss functions and the model's weight geometry, not of the specific
corpus.

\textbf{The dose-response curve as implicit control.} Although we do not
include an explicit $\lambda$\_MRP=0 (pure CE) control run, the dose-response
design itself isolates the MRP contribution. Cross-entropy weight is
held constant at 1.0 across all runs; only $\lambda$\_MRP varies. If CE were
responsible for the margin improvements, all runs would show the same
effect regardless of $\lambda$\_MRP. Instead, Fisher median margin scales
monotonically with $\lambda$\_MRP (1.174 at 0.15 $\rightarrow$ 1.317 at 0.6 $\rightarrow$ 1.630 at 2.0),
and the two loss functions produce qualitatively different damage
profiles at matched $\lambda$ values. Both observations can only be explained by
the MRP term.

The main fully preserved sweep tests:

\begin{itemize}
\tightlist
\item
  \textbf{Margin maximization:} $\lambda$\_MRP $\in$ \{0.1, 0.15, 0.3, 1.0\}
\item
  \textbf{Fisher information:} $\lambda$\_MRP $\in$ \{0.15, 0.3, 0.6\}
\end{itemize}

Each run is followed by a full fp32 margin audit and per-position
comparison to baseline, yielding 256,577-position churn analysis.

Two additional higher-$\lambda$ Fisher runs extend the dose-response curve:

\begin{itemize}
\tightlist
\item
  \textbf{Fisher information, extended range:} $\lambda$\_MRP $\in$ \{1.0, 2.0\}
\end{itemize}

For these points, geometry audits, benchmark evaluations, and
per-position audit data are available, enabling churn, frequency, and
token-class analyses. These higher-$\lambda$ runs do not address the
domain-specific token-value question.

\hypertarget{downstream-evaluation}{%
\subsection{Downstream Evaluation}\label{downstream-evaluation}}

Six loglikelihood-only tasks from lm-evaluation-harness: ARC-Challenge,
HellaSwag, WinoGrande, PIQA, LAMBADA (OpenAI), TruthfulQA MC1.

\hypertarget{parameter-update-scope}{%
\subsection{Parameter Update Scope}\label{parameter-update-scope}}

All MRP experiments use full-model training scope
(\texttt{-\/-trainable-scope\ text}): all transformer body parameters
plus the \texttt{lm\_head} projection are updated. Qwen3.5-4B uses tied
input/output embeddings (\texttt{tie\_word\_embeddings=True}), meaning
\texttt{lm\_head.weight} and \texttt{model.embed\_tokens.weight} share
the same tensor. Updates to the lm\_head therefore simultaneously modify
all token input representations --- the observed effects are not
isolated to output-layer Voronoi boundary reshaping but include changes
to the full representation pipeline.

We chose full-model scope for the initial dose-response study because it
maximizes the optimizer's freedom to redistribute margin across the
manifold. However, this makes it difficult to isolate whether the
observed Fisher behavior arises specifically from Voronoi boundary
reshaping or from broader representational changes. Ablations with
frozen backbone (e.g., updating only the final norm and/or lm\_head, or
using low-rank adapters) would help isolate the minimal parameter subset
responsible for the geometric effects and are an important direction for
future work.

\hypertarget{gradient-treatment-of-topk-selection}{%
\subsection{Gradient Treatment of TopK
Selection}\label{gradient-treatment-of-topk-selection}}

Both the margin loss (Algorithm 1) and Fisher loss (Algorithm 2) use
hard TopK selection (\texttt{torch.topk}). The index selection is
non-differentiable: gradients flow through the selected logit
\emph{values} and embedding \emph{rows} but not through the discrete
choice of \emph{which} tokens are selected. There is no straight-through
estimator or soft relaxation.

For margin maximization (k=2), the gradient pushes the top-1 logit up
and the top-2 logit down at positions selected by the margin threshold
gate --- a hard Boolean mask that is also non-differentiable.

For Fisher MRP (k=5), the gradient flows through: (1) the softmax
probabilities over the top-k logits, (2) the L2-normalized embedding
rows gathered at the top-k indices, and (3) the quadratic form computing
pairwise Fisher distances. Embedding rows outside the top-k receive zero
gradient from the Fisher loss. The top-k set is fixed per forward pass:
a token ranked 6th receives no Fisher gradient regardless of its
proximity to rank 5.

We expect this hard selection to be stable in practice because: (a) at
k=5, the competitive token set changes infrequently between consecutive
gradient steps for a well-trained model; (b) the probability weighting
\texttt{p(t\_i)\ $\cdot$\ p(t\_j)} in the Fisher loss naturally downweights
tokens near the selection boundary (rank 5 carries low probability
mass); and (c) the \texttt{sqrt(clamp(...,\ min=1e-8))} in the distance
computation prevents gradient divergence at near-zero distances.
Empirically, the loss curves are smooth across all runs with no
instability artifacts, but we have not systematically measured top-k set
stability across steps or sensitivity to k.

\hypertarget{results}{%
\section{Results}\label{results}}

\hypertarget{expressibility-gap-scaling}{%
\subsection{Expressibility Gap
Scaling}\label{expressibility-gap-scaling}}

The log-log regression of $\hat{\eta}$($\varepsilon$) on the float32 margins yields:

\begin{longtable}[]{@{}llll@{}}
\toprule\noalign{}
Metric & bf16 (original) & fp32 (recomputed) & Mabrok range \\
\midrule\noalign{}
\endhead
\bottomrule\noalign{}
\endlastfoot
$\beta$ (log-log slope) & 0.957 & 0.912 & 0.873-1.117 \\
R$^2$ & 0.927 & 0.9997 & \textgreater{} 0.985 \\
$\alpha$ (gap coefficient) & 0.655 & 0.762 & not reported \\
\end{longtable}

R$^2$ = 0.9997 is the highest reported for any model on this test. The
slope $\beta$ = 0.912 falls within Mabrok's observed range {[}0.873, 1.117{]},
confirming Theorem 10.5 on a Gated DeltaNet architecture with 248K
vocabulary. The sub-unity slope is expected: Theorem 2.3 predicts $\eta$($\varepsilon$) =
$\alpha$$\cdot$$\varepsilon$ + O($\varepsilon$$^2$), and the positive higher-order correction at the finite $\varepsilon$
values used in the regression pulls the apparent log-log slope below 1.
All six of Mabrok's architectures exhibited the same effect.

\hypertarget{margin-distribution}{%
\subsection{Margin Distribution}\label{margin-distribution}}

\begin{longtable}[]{@{}lll@{}}
\toprule\noalign{}
Metric & Qwen3.5-4B & Mabrok range (124M-1.5B) \\
\midrule\noalign{}
\endhead
\bottomrule\noalign{}
\endlastfoot
m$_0$.$_0$$_5$ & 0.065 & 0.041-0.055 \\
m$_0$.$_2$$_5$ & 0.401 & 0.228-0.297 \\
Median & 1.030 & 0.487-0.641 \\
m$_0$.$_7$$_5$ & 2.287 & 0.856-1.055 \\
m$_0$.$_9$$_5$ & 6.153 & 1.484-1.756 \\
Pr(m \textless{} 0.5) & 0.300 & 0.395-0.512 \\
\end{longtable}

Margins approximately 2x those of Mabrok's largest model (OPT-1.3B),
consistent with improved tessellation quality at the 4B scale. The
ambiguity floor m$_0$.$_0$$_5$ = 0.065 sits slightly above Mabrok's observed
range (0.041--0.055), suggesting that the 4B model partially resolves
some positions that smaller models treat as irreducibly ambiguous.

\hypertarget{layer-wise-mrp-ce-correlation-and-mid-layer-geometric-ambiguity}{%
\subsection{Layer-wise MRP-CE Correlation and Mid-Layer Geometric
Ambiguity}\label{layer-wise-mrp-ce-correlation-and-mid-layer-geometric-ambiguity}}

We project each layer's hidden state through the final lm\_head to
compute ``virtual'' margins and MRP penalties at every depth. These
virtual logits measure the mismatch between intermediate representation
geometry and final linear decodability; they should not be interpreted
as a native intermediate classifier.

Three regimes emerge:

\begin{enumerate}
\def\labelenumi{\arabic{enumi}.}
\item
  \textbf{Layers 4-20 (contextual integration):} Voronoi geometry is
  meaningless. Margins near zero, MRP uniformly high, no correlation
  with final-layer CE.
\item
  \textbf{Layers 24-28 (mid-layer geometric ambiguity):} CE-MRP
  correlation is \emph{negative} (Spearman $\rho$ = -0.29, n = 256,577
  positions, p \textless{} 10$^{-5}$). When intermediate hidden states are
  projected through the final lm\_head, positions that are ultimately
  predicted correctly often remain geometrically ambiguous at these
  depths. Late layers then sharpen and linearize these states into the
  final correct decision.
\item
  \textbf{Layers 31-32 (prediction crystallization):} CE-MRP correlation
  is strongly positive (Spearman $\rho$ = 0.836, n = 256,577, p \textless{}
  10$^{-5}$). Voronoi geometry aligns with prediction quality. MRP is largely
  redundant with CE.
\end{enumerate}

All correlations in this paper are Spearman rank correlations computed
over the full 256,577-position evaluation set unless otherwise noted. At
this sample size, all reported correlations with \textbar $\rho$\textbar{}
\textgreater{} 0.01 are significant at p \textless{} 10$^{-5}$.

\hypertarget{dose-response-margin-statistics}{%
\subsection{Dose-Response: Margin
Statistics}\label{dose-response-margin-statistics}}

\begin{longtable}[]{@{}lllllll@{}}
\toprule\noalign{}
Model & Median & $\Delta$ Median & Pr(m\textless0.5) & $\Delta$ Pr & $\alpha$ & R$^2$ \\
\midrule\noalign{}
\endhead
\bottomrule\noalign{}
\endlastfoot
Baseline & 1.030 & --- & 0.300 & --- & 0.762 & 0.9997 \\
MM $\lambda$\_MRP=0.1 & 1.288 & +25.0\% & 0.236 & -21.3\% & 0.535 & 0.9995 \\
MM $\lambda$\_MRP=0.15 & 1.366 & +32.6\% & 0.222 & -26.0\% & 0.494 & 0.9995 \\
MM $\lambda$\_MRP=0.3 & 1.571 & +52.5\% & 0.180 & -39.9\% & 0.389 & 0.9995 \\
MM $\lambda$\_MRP=1.0 & 1.932 & +87.5\% & 0.062 & -79.5\% & 0.130 & 0.9881 \\
Fish $\lambda$\_MRP=0.15 & 1.174 & +13.9\% & 0.271 & -9.7\% & 0.670 & 0.9994 \\
Fish $\lambda$\_MRP=0.3 & 1.220 & +18.4\% & 0.263 & -12.2\% & 0.649 & 0.9995 \\
Fish $\lambda$\_MRP=0.6 & 1.317 & +27.8\% & 0.250 & -16.7\% & 0.612 & 0.9996 \\
Fish $\lambda$\_MRP=1.0 & 1.350 & +31.1\% & 0.244 & -18.6\% & 0.599 & 0.9995 \\
Fish $\lambda$\_MRP=2.0 & 1.630 & +58.3\% & 0.214 & -28.6\% & 0.514 & 0.9996 \\
\end{longtable}

Both methods reduce the expressibility gap monotonically with $\lambda$\_MRP.
The power-law fit quality (R$^2$) is preserved at $\geq$0.9994 for all runs
except MM $\lambda$\_MRP=1.0 (R$^2$ = 0.9881), indicating the scaling law holds
under margin refinement. The 1.0/2.0 Fisher rows extend the geometry,
churn, per-band, benchmark, and coarse token-value analysis.

\hypertarget{dose-response-prediction-churn}{%
\subsection{Dose-Response: Prediction
Churn}\label{dose-response-prediction-churn}}

\begin{longtable}[]{@{}llllll@{}}
\toprule\noalign{}
Model & Churn\% & W$\rightarrow$R & R$\rightarrow$W & Flip Ratio & Net Corrected \\
\midrule\noalign{}
\endhead
\bottomrule\noalign{}
\endlastfoot
MM $\lambda$\_MRP=0.1 & 20.0\% & 16,358 & 5,716 & 2.9x & +10,642 \\
MM $\lambda$\_MRP=0.15 & 20.5\% & 16,390 & 6,065 & 2.7x & +10,325 \\
MM $\lambda$\_MRP=0.3 & 22.9\% & 16,680 & 7,910 & 2.1x & +8,770 \\
MM $\lambda$\_MRP=1.0 & 50.8\% & 16,719 & 39,108 & 0.4x & -22,389 \\
Fish $\lambda$\_MRP=0.15 & 19.3\% & 16,195 & 5,317 & 3.0x & +10,878 \\
Fish $\lambda$\_MRP=0.3 & 19.3\% & 16,256 & 5,298 & 3.1x & +10,958 \\
Fish $\lambda$\_MRP=0.6 & 19.4\% & 16,327 & 5,356 & 3.0x & +10,971 \\
Fish $\lambda$\_MRP=1.0 & 18.0\% & 14,912 & 4,994 & 3.0x & +9,918 \\
Fish $\lambda$\_MRP=2.0 & 19.3\% & 15,350 & 5,842 & 2.6x & +9,508 \\
\end{longtable}

\textbf{Finding 1: Corrections ceiling.} W$\rightarrow$R (wrong $\rightarrow$ right) corrections
plateau near \textasciitilde16,300 across margin max and Fisher through
$\lambda$\_MRP=0.6, then soften modestly at higher Fisher strengths (14,912 at
1.0, 15,350 at 2.0). This still supports the idea of a largely fixed
``geometrically accessible'' correction reservoir, but it also suggests
that very high Fisher strength starts reallocating some of that
accessible capacity toward more concentrated head-token cleanup rather
than maximizing total corrections.

\textbf{Finding 2: Fisher damage is flat through $\lambda$\_MRP=0.6 and still
much flatter than margin max beyond it.} Fisher R$\rightarrow$W (right $\rightarrow$ wrong)
damage is 5,317/5,298/5,356 across $\lambda$\_MRP = 0.15/0.3/0.6 ---
near-constant across dose levels (range 5,298--5,356, span
\textless0.6\% of mean). At 1.0, damage remains low (4,994) and the flip
ratio stays near 3.0x; by 2.0, damage rises to 5,842 and the flip ratio
declines to 2.6x, still far from the catastrophic margin-max $\lambda$\_MRP=1.0
regime.

\textbf{Finding 3: Fisher's geometry continues improving past
$\lambda$\_MRP=0.6, but the practical tradeoff changes.} The $\lambda$\_MRP = 1.0 and
2.0 runs continue improving median margin and Pr(m\textless0.5), and
benchmark means remain within the same narrow band as baseline. However,
frequency and token-class analyses show that the higher-$\lambda$ gains are
increasingly concentrated in head and structural tokens rather than
uniformly distributed across the vocabulary.

\hypertarget{runner-up-rotation-analysis}{%
\subsection{Runner-Up Rotation
Analysis}\label{runner-up-rotation-analysis}}

Approximately 25\% of positions retain their top-1 prediction but change
their runner-up token after MRP. We call this ``runner-up rotation.''

\begin{longtable}[]{@{}llll@{}}
\toprule\noalign{}
Model & Rotated\% & Wider Margin\% & Mean Margin $\Delta$ \\
\midrule\noalign{}
\endhead
\bottomrule\noalign{}
\endlastfoot
MM $\lambda$\_MRP=0.1 & 25.1\% & 68.6\% & +0.820 \\
MM $\lambda$\_MRP=0.3 & 26.2\% & 74.8\% & +0.902 \\
Fish $\lambda$\_MRP=0.15 & 24.7\% & 64.8\% & +0.790 \\
Fish $\lambda$\_MRP=0.3 & 24.8\% & 67.5\% & +0.840 \\
Fish $\lambda$\_MRP=0.6 & 24.8\% & 72.0\% & +0.922 \\
Fish $\lambda$\_MRP=1.0 & 23.0\% & 75.0\% & +0.949 \\
Fish $\lambda$\_MRP=2.0 & 24.1\% & 80.0\% & +1.149 \\
\end{longtable}

Fisher $\lambda$\_MRP=0.6 already matches margin max $\lambda$\_MRP=0.3 on rotation
margin delta (+0.922), and the 1.0/2.0 runs continue the same widening
trend (+0.949, +1.149). Fisher rotates the runner-up to a different,
more naturally separable token --- the model finds an alternative
second-place prediction that the representation space can accommodate at
greater distance.

\hypertarget{accuracy-by-margin-band}{%
\subsection{Accuracy by Margin Band}\label{accuracy-by-margin-band}}

\begin{longtable}[]{@{}
  >{\raggedright\arraybackslash}p{(\columnwidth - 12\tabcolsep) * \real{0.1591}}
  >{\raggedright\arraybackslash}p{(\columnwidth - 12\tabcolsep) * \real{0.1591}}
  >{\raggedright\arraybackslash}p{(\columnwidth - 12\tabcolsep) * \real{0.1591}}
  >{\raggedright\arraybackslash}p{(\columnwidth - 12\tabcolsep) * \real{0.1136}}
  >{\raggedright\arraybackslash}p{(\columnwidth - 12\tabcolsep) * \real{0.1136}}
  >{\raggedright\arraybackslash}p{(\columnwidth - 12\tabcolsep) * \real{0.0909}}
  >{\raggedright\arraybackslash}p{(\columnwidth - 12\tabcolsep) * \real{0.2045}}@{}}
\toprule\noalign{}
\begin{minipage}[b]{\linewidth}\raggedright
Model
\end{minipage} & \begin{minipage}[b]{\linewidth}\raggedright
m\textless0.5
\end{minipage} & \begin{minipage}[b]{\linewidth}\raggedright
0.5-1
\end{minipage} & \begin{minipage}[b]{\linewidth}\raggedright
1-2
\end{minipage} & \begin{minipage}[b]{\linewidth}\raggedright
2-5
\end{minipage} & \begin{minipage}[b]{\linewidth}\raggedright
5+
\end{minipage} & \begin{minipage}[b]{\linewidth}\raggedright
Overall
\end{minipage} \\
\midrule\noalign{}
\endhead
\bottomrule\noalign{}
\endlastfoot
Baseline & 21.6\% (77K) & 31.5\% (49K) & 46.6\% (56K) & 78.1\% (53K) &
99.3\% (21K) & 47.1\% \\
MM $\lambda$\_MRP=0.3 & 21.5\% (46K) & 27.4\% (38K) & 37.5\% (69K) & 72.1\%
(71K) & 99.3\% (33K) & 50.5\% \\
Fish $\lambda$\_MRP=0.3 & 22.6\% (68K) & 31.6\% (46K) & 46.6\% (54K) & 78.3\%
(56K) & 99.4\% (33K) & 51.4\% \\
Fish $\lambda$\_MRP=0.6 & 22.1\% (64K) & 30.9\% (44K) & 44.6\% (54K) & 75.9\%
(60K) & 99.3\% (35K) & 51.4\% \\
Fish $\lambda$\_MRP=1.0 & 22.0\% (63K) & 30.3\% (43K) & 43.3\% (54K) & 74.1\%
(61K) & 99.1\% (35K) & 51.0\% \\
Fish $\lambda$\_MRP=2.0 & 21.2\% (55K) & 28.7\% (38K) & 39.3\% (52K) & 67.2\%
(72K) & 98.4\% (40K) & 50.8\% \\
\end{longtable}

Fisher preserves per-band accuracy well through $\lambda$\_MRP=0.6 while
shifting positions to higher bands. At MM $\lambda$\_MRP=0.3, mid-band accuracy
degrades (37.5\% at 1-2 band vs baseline 46.6\%). Fisher $\lambda$\_MRP=0.3
maintains identical accuracy at 1-2 (46.6\%) with 2K fewer positions,
indicating clean band transitions rather than forced displacement. The
1.0/2.0 runs show that the band shift continues, but per-band accuracy
softens beyond 0.6, especially in the 1--2 and 2--5 ranges.

\hypertarget{margin-expansion-distribution}{%
\subsection{Margin Expansion
Distribution}\label{margin-expansion-distribution}}

\begin{longtable}[]{@{}llll@{}}
\toprule\noalign{}
Model & Positions Wider & Mean $\Delta$ & Median $\Delta$ \\
\midrule\noalign{}
\endhead
\bottomrule\noalign{}
\endlastfoot
MM $\lambda$\_MRP=0.1 & 65.4\% & +0.507 & +0.197 \\
MM $\lambda$\_MRP=0.3 & 72.8\% & +0.653 & +0.408 \\
Fish $\lambda$\_MRP=0.15 & 60.0\% & +0.448 & +0.112 \\
Fish $\lambda$\_MRP=0.3 & 63.1\% & +0.493 & +0.151 \\
Fish $\lambda$\_MRP=0.6 & 68.2\% & +0.574 & +0.232 \\
\end{longtable}

\hypertarget{downstream-benchmarks}{%
\subsection{Downstream Benchmarks}\label{downstream-benchmarks}}

\begin{longtable}[]{@{}llllllll@{}}
\toprule\noalign{}
Model & ARC-C & HellaSwag & WinoGrande & PIQA & LAMBADA & TQA & Mean \\
\midrule\noalign{}
\endhead
\bottomrule\noalign{}
\endlastfoot
Baseline & .528 & .751 & .709 & .784 & .668 & .354 & .632 \\
MM $\lambda$\_MRP=0.15 & .535 & .754 & .706 & .785 & .683 & .355 & .636 \\
MM $\lambda$\_MRP=0.3 & .529 & .754 & .696 & .786 & .698 & .353 & .636 \\
Fish $\lambda$\_MRP=0.15 & .534 & .754 & .700 & .785 & .670 & .351 & .632 \\
Fish $\lambda$\_MRP=0.3 & .535 & .754 & .706 & .784 & .671 & .349 & .633 \\
Fish $\lambda$\_MRP=0.6 & .527 & .751 & .710 & .783 & .667 & .354 & .632 \\
Fish $\lambda$\_MRP=1.0 & .529 & .756 & .709 & .787 & .677 & .346 & .634 \\
Fish $\lambda$\_MRP=2.0 & .529 & .752 & .704 & .785 & .674 & .343 & .631 \\
\end{longtable}

Maximum deviation from baseline mean across all runs is small (means
remain in the 0.631--0.636 band). Task means remain broadly flat despite
substantial token-level boundary reorganization. This is consistent with
gains concentrated in head and structural tokens, but does not by itself
identify which token classes are benefiting. The remaining question at
higher $\lambda$\_MRP is therefore not benchmark recovery but whether
increasingly concentrated structural/head-token gains represent an
acceptable tradeoff.

\hypertarget{qwen-frequency-stratified-flip-audit}{%
\subsection{Qwen Frequency-Stratified Flip
Audit}\label{qwen-frequency-stratified-flip-audit}}

To test whether the frequency concentration observed at 264M scale
generalizes, we bucketed the 256,577 audited positions by target-token
frequency in WikiText validation and compared the baseline audit against
Fisher-polished audits at $\lambda$\_MRP = 0.6, 1.0, and 2.0.

\begin{longtable}[]{@{}
  >{\raggedright\arraybackslash}p{(\columnwidth - 14\tabcolsep) * \real{0.1212}}
  >{\raggedright\arraybackslash}p{(\columnwidth - 14\tabcolsep) * \real{0.1515}}
  >{\raggedright\arraybackslash}p{(\columnwidth - 14\tabcolsep) * \real{0.1515}}
  >{\raggedright\arraybackslash}p{(\columnwidth - 14\tabcolsep) * \real{0.0909}}
  >{\raggedright\arraybackslash}p{(\columnwidth - 14\tabcolsep) * \real{0.1515}}
  >{\raggedright\arraybackslash}p{(\columnwidth - 14\tabcolsep) * \real{0.0909}}
  >{\raggedright\arraybackslash}p{(\columnwidth - 14\tabcolsep) * \real{0.1515}}
  >{\raggedright\arraybackslash}p{(\columnwidth - 14\tabcolsep) * \real{0.0909}}@{}}
\toprule\noalign{}
\begin{minipage}[b]{\linewidth}\raggedright
Bucket
\end{minipage} & \begin{minipage}[b]{\linewidth}\raggedright
Baseline
\end{minipage} & \begin{minipage}[b]{\linewidth}\raggedright
Fish 0.6
\end{minipage} & \begin{minipage}[b]{\linewidth}\raggedright
$\Delta$0.6
\end{minipage} & \begin{minipage}[b]{\linewidth}\raggedright
Fish 1.0
\end{minipage} & \begin{minipage}[b]{\linewidth}\raggedright
$\Delta$1.0
\end{minipage} & \begin{minipage}[b]{\linewidth}\raggedright
Fish 2.0
\end{minipage} & \begin{minipage}[b]{\linewidth}\raggedright
$\Delta$2.0
\end{minipage} \\
\midrule\noalign{}
\endhead
\bottomrule\noalign{}
\endlastfoot
1 & 0.283 & 0.294 & +0.011 & 0.287 & +0.004 & 0.284 & +0.001 \\
2-4 & 0.310 & 0.324 & +0.014 & 0.320 & +0.011 & 0.318 & +0.008 \\
5-19 & 0.341 & 0.356 & +0.015 & 0.351 & +0.010 & 0.347 & +0.006 \\
20-99 & 0.391 & 0.409 & +0.019 & 0.404 & +0.014 & 0.399 & +0.008 \\
100+ & 0.557 & 0.620 & +0.063 & 0.616 & +0.059 & 0.617 & +0.059 \\
\end{longtable}

The gains are positive in every bucket at $\lambda$\_MRP=0.6, but heavily
concentrated in the highest-frequency bucket. Net corrected positions
from the 100+ bucket account for 84.0\% of the total Fisher gain at 0.6,
87.5\% at 1.0, and 92.1\% at 2.0. The concentration becomes stronger as
$\lambda$\_MRP increases even though benchmark means remain flat.

The top gains are dominated by structural tokens and formatting pieces:
commas, periods, spaces, quotes, hyphens, ``@'', and ``=''-like
separators. The most persistent negative-net regressions at higher
$\lambda$\_MRP are not exotic rare fragments but common functional/content
tokens such as:

\emph{and}, \emph{he}, \emph{his}, \emph{with}, \emph{which}, \emph{it},
\emph{they}, and \emph{their}.

Fisher is not purely a punctuation optimizer, but its largest gains come
from head tokens, and its higher-$\lambda$ regressions cluster in ordinary
medium-frequency words.

\hypertarget{qwen-heuristic-token-class-audit}{%
\subsection{Qwen Heuristic Token-Class
Audit}\label{qwen-heuristic-token-class-audit}}

To go beyond raw frequency, we replayed the same per-position audits
with heuristic token-class rules applied to the decoded token text. The
classification hierarchy (applied in order, first match wins):

\begin{enumerate}
\def\labelenumi{\arabic{enumi}.}
\tightlist
\item
  \textbf{Structural:} all characters are Unicode punctuation (category
  P*) or symbols (category S*), or the token is whitespace-only.
  Examples: \texttt{,}, \texttt{.}, \texttt{"}, \texttt{@}, \texttt{=},
  \texttt{---}.
\item
  \textbf{Numeric:} matches \texttt{\^{}{[}0-9{]}{[}0-9,./:\%+-{]}*\$}.
  Examples: \texttt{123}, \texttt{2023}, \texttt{3.14}, \texttt{50\%}.
\item
  \textbf{Function word:} pure alphabetic token whose lowercased form
  appears in a fixed 101-word set of articles, prepositions, pronouns,
  auxiliaries, and conjunctions (e.g., \emph{the}, \emph{of}, \emph{is},
  \emph{they}, \emph{which}).
\item
  \textbf{Entity-like:} starts with a capital letter followed by
  lowercase letters (\texttt{\^{}{[}A-Z{]}{[}A-Za-z{]}+\$}) or is
  all-caps with 2+ letters (\texttt{\^{}{[}A-Z{]}\{2,\}\$}). Examples:
  \emph{Paris}, \emph{John}, \emph{USA}, \emph{NLP}.
\item
  \textbf{Content word:} remaining pure alphabetic tokens (nouns, verbs,
  adjectives, adverbs not in the function-word set).
\item
  \textbf{Fragment/other:} everything else (mixed alphanumeric, subword
  fragments with special characters).
\end{enumerate}

The full classification code and word list are in
\texttt{scripts/analysis/token\_class\_flip\_audit.py}. This is a coarse
taxonomy rather than a linguistic gold standard, but it distinguishes
punctuation/formatting cleanup from content-bearing gains. Approximately
33 positions per run (\textasciitilde0.3\%) fall into the fragment/other
class and are omitted from the table below.

\begin{longtable}[]{@{}
  >{\raggedright\arraybackslash}p{(\columnwidth - 12\tabcolsep) * \real{0.1279}}
  >{\raggedright\arraybackslash}p{(\columnwidth - 12\tabcolsep) * \real{0.1628}}
  >{\raggedright\arraybackslash}p{(\columnwidth - 12\tabcolsep) * \real{0.1279}}
  >{\raggedright\arraybackslash}p{(\columnwidth - 12\tabcolsep) * \real{0.1628}}
  >{\raggedright\arraybackslash}p{(\columnwidth - 12\tabcolsep) * \real{0.1279}}
  >{\raggedright\arraybackslash}p{(\columnwidth - 12\tabcolsep) * \real{0.1628}}
  >{\raggedright\arraybackslash}p{(\columnwidth - 12\tabcolsep) * \real{0.1279}}@{}}
\toprule\noalign{}
\begin{minipage}[b]{\linewidth}\raggedright
Class
\end{minipage} & \begin{minipage}[b]{\linewidth}\raggedright
Fish 0.6 Net
\end{minipage} & \begin{minipage}[b]{\linewidth}\raggedright
Share 0.6
\end{minipage} & \begin{minipage}[b]{\linewidth}\raggedright
Fish 1.0 Net
\end{minipage} & \begin{minipage}[b]{\linewidth}\raggedright
Share 1.0
\end{minipage} & \begin{minipage}[b]{\linewidth}\raggedright
Fish 2.0 Net
\end{minipage} & \begin{minipage}[b]{\linewidth}\raggedright
Share 2.0
\end{minipage} \\
\midrule\noalign{}
\endhead
\bottomrule\noalign{}
\endlastfoot
Structural & 7,231 & 65.9\% & 7,001 & 70.6\% & 7,135 & 75.0\% \\
Function & 1,208 & 11.0\% & 979 & 9.9\% & 982 & 10.3\% \\
Content & 1,708 & 15.6\% & 1,284 & 13.0\% & 890 & 9.4\% \\
Entity-like & 356 & 3.2\% & 240 & 2.4\% & 82 & 0.9\% \\
Numeric & 435 & 4.0\% & 382 & 3.9\% & 386 & 4.1\% \\
\end{longtable}

Shares are computed against the full net corrected count from Table 4.5.

The token-class view sharpens the frequency result. At $\lambda$\_MRP=0.6,
Fisher produces positive net corrections for every coarse class,
including content words and entity-like tokens, but the gains are
already dominated by structural cleanup. As $\lambda$\_MRP rises, structural
share increases while content-word and entity-like contributions shrink.
By $\lambda$\_MRP=2.0, entity-like tokens are barely net positive and content
words contribute less than ten percent of the total gain.

\hypertarget{fine-tuned-instruct-sample}{%
\subsection{Fine-Tuned Instruct
Sample}\label{fine-tuned-instruct-sample}}

To test whether these effects survive fine-tuning, we analyzed a single
fine-tuned Qwen3.5-4B instruct-style model. This is a single sample, not
a matched instruct-model sweep, so it should be read as an initial probe
rather than a full replication.

\begin{longtable}[]{@{}
  >{\raggedright\arraybackslash}p{(\columnwidth - 6\tabcolsep) * \real{0.2121}}
  >{\raggedright\arraybackslash}p{(\columnwidth - 6\tabcolsep) * \real{0.2879}}
  >{\raggedright\arraybackslash}p{(\columnwidth - 6\tabcolsep) * \real{0.2576}}
  >{\raggedright\arraybackslash}p{(\columnwidth - 6\tabcolsep) * \real{0.2424}}@{}}
\toprule\noalign{}
\begin{minipage}[b]{\linewidth}\raggedright
Metric
\end{minipage} & \begin{minipage}[b]{\linewidth}\raggedright
Instruct Baseline
\end{minipage} & \begin{minipage}[b]{\linewidth}\raggedright
Instruct Fisher
\end{minipage} & \begin{minipage}[b]{\linewidth}\raggedright
$\Delta$ / Note
\end{minipage} \\
\midrule\noalign{}
\endhead
\bottomrule\noalign{}
\endlastfoot
Token accuracy & 45.4\% & 48.1\% & +2.63\% \\
Flip ratio & --- & 1.74x & materially weaker than base-model Fisher \\
Head-bucket share & --- & 85.0\% & 100+ frequency bucket share of net
gain \\
Structural class share & --- & 58.8\% & dominant source of gains \\
Entity-like net & --- & -219 & net negative \\
6-task mean* & --- & 0.632 & no matched baseline bundle preserved \\
\end{longtable}

*Post-run benchmark mean over ARC-C, HellaSwag, WinoGrande, PIQA,
LAMBADA, and TruthfulQA MC1.

The instruct result is directionally consistent with the base-model
findings. Head-token concentration survives fine-tuning, and the
practical tradeoff may be harsher: the overall flip ratio is lower, and
the entity-like class is net negative even though structural and
function-word cleanup remain positive. This does not prove that
fine-tuning always worsens Fisher's token-value profile, but it cautions
that base-model gains should not be assumed to transfer cleanly through
alignment or supervised tuning.

\hypertarget{preliminary-targeted-and-protected-variants}{%
\subsection{Preliminary Targeted and Protected
Variants}\label{preliminary-targeted-and-protected-variants}}

To explore whether Fisher can be turned from a bulk correction
instrument into a more surgical editing tool, we evaluated targeted and
protected variants. These runs are heterogeneous and should be read as
pilot evidence rather than a matched sweep. The goal is not to claim a
solved protection recipe, but to test whether targeting or protection
already improves the token-value profile.

\begin{longtable}[]{@{}lllll@{}}
\toprule\noalign{}
Variant & $\Delta$ Acc & Flip Ratio & Head Share of Net & Entity-like Net \\
\midrule\noalign{}
\endhead
\bottomrule\noalign{}
\endlastfoot
Bulk Fish $\lambda$\_MRP=0.6 & +4.28\% & 3.05x & 84.0\% & +356 \\
Targeted base fix03 & +2.02\% & 1.94x & 94.1\% & -61 \\
Targeted base fix06 & +1.95\% & 1.86x & 94.9\% & -105 \\
Targeted instruct fix03 & +2.58\% & 1.68x & 83.4\% & -98 \\
Targeted instruct v2 fix03 & +2.63\% & 1.74x & 85.0\% & -219 \\
\end{longtable}

The result is mixed. Targeted Fisher variants remain net positive, but
on the base model they are even more head-concentrated than bulk Fisher
$\lambda$\_MRP=0.6 and already flip the coarse entity-like class negative. The
instruct-style targeted variants show the same pattern after
fine-tuning: aggregate token accuracy still rises, yet entity-like
tokens remain net negative and the flip ratio is weaker than the
base-model bulk Fisher result. The current targeted recipe does not yet
demonstrate the desired ``high-value tail preserved, head cleaned up''
behavior.

We also examined 264M-scale protection pilots using mean-margin
protection and Elastic Weight Consolidation (EWC; Kirkpatrick et al.,
2017):

\begin{longtable}[]{@{}lllll@{}}
\toprule\noalign{}
264M Pilot & Accuracy & Flip Ratio & W$\rightarrow$R & R$\rightarrow$W \\
\midrule\noalign{}
\endhead
\bottomrule\noalign{}
\endlastfoot
Mean-protect, 50 steps & 0.3938 & 2.90x & 255 & 88 \\
Mean-protect, 200 steps & 0.4013 & 2.43x & 336 & 138 \\
Mean-protect + EWC $\lambda$=1 & 0.4003 & 2.42x & 331 & 137 \\
Mean-protect + EWC $\lambda$=10000 & 0.4013 & 2.43x & 336 & 138 \\
\end{longtable}

These pilots suggest that shorter mean-protect passes may be somewhat
safer than longer ones, but they do not show a clear EWC benefit.

\hypertarget{discussion-and-future-exploration}{%
\section{Discussion and Future
Exploration}\label{discussion-and-future-exploration}}

\hypertarget{mrp-does-work}{%
\subsection{MRP Does Work}\label{mrp-does-work}}

Our initial probe in bf16 reported negative results from 100-500 step
MRP training. Those results do not accurately describe the effect of
margin refinement in either Margin Max or Fisher formulations, so they
have been omitted. The dose-response study reveals this was a
formulation problem, not a fundamental limitation. With direct margin
maximization or Fisher information distance as the loss, 200 steps
suffices to substantially reshape the Voronoi tessellation.

\hypertarget{the-corrections-ceiling}{%
\subsection{The Corrections Ceiling}\label{the-corrections-ceiling}}

The W$\rightarrow$R correction count is near-invariant through the main validated
range: roughly \textasciitilde16,300 out of 256,577 positions across
margin max and Fisher up to $\lambda$\_MRP=0.6. At higher Fisher strengths it
softens modestly (14,912 at 1.0, 15,350 at 2.0), consistent with part of
the accessible reservoir being traded away as the objective becomes more
concentrated in head-token space. The ceiling interpretation still
holds: only a small fraction of positions are close enough to a better
Voronoi boundary that geometry alone can fix them. The remaining
incorrect positions are incorrect for reasons that geometry alone cannot
address --- missing knowledge, not boundary proximity.

This ceiling is a fundamental property of the model's representation
quality, not of the MRP method. It establishes that post-hoc margin
refinement can improve accuracy by at most \textasciitilde6.4 percentage
points on this model/corpus pair.

\hypertarget{why-fisher-damage-is-constant}{%
\subsection{Why Fisher Damage is
Constant}\label{why-fisher-damage-is-constant}}

The critical difference between margin maximization and Fisher
information distance is the direction of gradient pressure:

\textbf{Margin maximization} pushes along the fixed axis between the
current top-1 and top-2 logits. This is efficient when the two tokens
are genuinely confusable (similar semantics, similar embeddings), but
creates geometric distortion when they are not --- forcing separation
between tokens that the representation space places near each other for
good reason.

\textbf{Fisher information distance} moves along the direction of
maximum information divergence in the model's own metric tensor. When
Fisher MRP increases the margin at a position, it does so by rotating
the runner-up to a different token that is more naturally separable ---
one that the Fisher geometry identifies as the most ``different''
alternative. The top-1 prediction stays the same, but the second-place
alternative changes to one that the representation manifold can
accommodate at greater distance without distortion.

This explains the empirical observations:

\begin{enumerate}
\def\labelenumi{\arabic{enumi}.}
\tightlist
\item
  \textbf{Constant damage:} The rotation follows the model's natural
  geometry, so it doesn't create distortion regardless of magnitude.
\item
  \textbf{Increasing margin:} Higher $\lambda$\_MRP means further rotation along
  the same natural direction, not harder pushing along an unnatural one.
\item
  \textbf{Preserved per-band accuracy:} Band transitions are ``clean''
  --- positions move to higher bands because their geometry genuinely
  supports wider margins after runner-up rotation, not because they were
  forced there.
\end{enumerate}

\hypertarget{revised-higher-ux3bb-reading}{%
\subsection{Revised Higher-$\lambda$
Reading}\label{revised-higher-ux3bb-reading}}

The main sweep (Section 3.4) covers Fisher $\lambda$\_MRP through 0.6. Two
additional runs at $\lambda$\_MRP = 1.0 and 2.0 extend the dose-response curve:

\begin{itemize}
\tightlist
\item
  The geometric trend continues past 0.6: median margin rises (1.317 $\rightarrow$
  1.350 $\rightarrow$ 1.630) and Pr(m\textless0.5) falls (0.250 $\rightarrow$ 0.244 $\rightarrow$ 0.214).
\item
  Benchmark means remain near-flat at $\lambda$\_MRP=1.0 and 2.0.
\item
  The stronger claim that damage stays flat and token-value remains safe
  is only fully supported through $\lambda$\_MRP=0.6.
\end{itemize}

Three conclusions follow:

\begin{enumerate}
\def\labelenumi{\arabic{enumi}.}
\tightlist
\item
  Fisher's geometric improvements continue beyond the primary sweep
  range.
\item
  Higher $\lambda$\_MRP does not collapse benchmarks or the churn profile, but
  it does make the token-value distribution less uniform.
\item
  The practical operating ceiling is determined less by aggregate damage
  and more by whether the added gains at high $\lambda$\_MRP justify the growing
  concentration in structural/head-token space.
\end{enumerate}

Fisher $\lambda$\_MRP is therefore not a free parameter. Geometry continued to
improve through $\lambda$\_MRP = 2.0, but the practical cost shifted from
obvious benchmark damage to a subtler redistribution of token-level
utility.

\hypertarget{implications-for-the-expressibility-gap}{%
\subsection{Implications for the Expressibility
Gap}\label{implications-for-the-expressibility-gap}}

MRP directly compresses the expressibility gap $\eta$ by widening margins:
Fisher $\lambda$\_MRP=0.6 reduces Pr(m \textless{} 0.5) from 0.300 to 0.250
(-16.7\%) and the gap coefficient $\alpha$ from 0.762 to 0.612 (-19.7\%).
Critically, the power-law scaling R$^2$ is preserved (0.9996 vs 0.9997),
indicating that MRP reshapes the gap distribution while maintaining its
functional form. The gap is compressed, not broken.

This confirms that the linear scaling law of Theorem 10.5 describes a
stable geometric structure that persists under margin refinement --- the
Voronoi tessellation is plastic (can be reshaped) but topologically
invariant (retains its scaling character).

\hypertarget{the-redundancy-resolution}{%
\subsection{The Redundancy Resolution}\label{the-redundancy-resolution}}

Section 4.3 established that MRP loss is redundant with CE at the final
layer ($\rho$ = 0.836). Yet MRP demonstrably improves margins. How?

The high value correlation means that CE and MRP agree on \emph{which}
positions are problematic --- but they disagree on \emph{what to do
about them}. CE optimizes the absolute logit of the correct token. MRP
optimizes the gap between the top two logits, regardless of which is
correct. The two losses therefore have different gradient directions in
parameter space even when their magnitudes correlate across positions.
MRP provides gradient signal at positions where CE is satisfied (correct
prediction, reasonable confidence) but the margin is narrow ---
positions where MRP has independent information.

Note that value correlation across positions ($\rho$ = 0.836) does not imply
gradient alignment --- two losses can agree on which positions are hard
while prescribing different parameter updates. The dose-response results
confirm that MRP's gradient carries information that CE's does not:
margins improve monotonically with $\lambda$\_MRP in a regime where CE alone
would have no reason to change them.

\hypertarget{vocabulary-value-caveat}{%
\subsection{Vocabulary-Value Caveat}\label{vocabulary-value-caveat}}

The Qwen-native frequency audit confirms the practical caveat first
observed at 264M scale. Fisher improves aggregate token accuracy, but
the gain is highly concentrated in the head of the frequency
distribution. At $\lambda$\_MRP=0.6, the 100+ bucket contributes 84.0\% of total
net corrected positions; at 1.0 and 2.0, that concentration rises to
87.5\% and 92.1\%.

The Qwen result is more nuanced than the 264M finding. At $\lambda$\_MRP=0.6,
every frequency bucket is still net positive, so Fisher is not purely a
punctuation optimizer. The precise characterization is:

\begin{itemize}
\tightlist
\item
  Fisher improves every frequency bucket at the main operating point
  $\lambda$\_MRP=0.6.
\item
  The effect size is overwhelmingly concentrated in very high-frequency
  tokens.
\item
  As $\lambda$\_MRP increases, that concentration becomes more extreme.
\item
  Benchmark means can remain flat even while token-value utility becomes
  less evenly distributed across the vocabulary.
\end{itemize}

The top token-level gains are dominated by punctuation and formatting
pieces such as commas, periods, spaces, quotes, hyphens, @, and =. The
dominant negative-net regressions at higher $\lambda$\_MRP are common
medium-frequency words: \emph{and}, \emph{he}, \emph{his}, \emph{with},
\emph{which}, \emph{it}, \emph{they}, and \emph{their}. The observed
failure mode is not long-tail damage but a redistribution of token-level
utility toward high-frequency structural tokens and away from
medium-frequency content/function words as Fisher strength rises.

Prior work has shown that maximum-likelihood language modeling is
frequency-biased and that performance on rare tokens is especially
fragile under token-frequency imbalance (Diehl Martinez et al., 2024;
Chung and Kim, 2025). Our result suggests that post-hoc Fisher
refinement inherits that bias and, at higher $\lambda$\_MRP, may amplify it in
ways not visible in flat benchmark means.

This does \textbf{not} overturn the geometric result. Fisher clearly
reshapes the margin distribution and improves aggregate token accuracy.
But it weakens the stronger practical claim that Fisher is automatically
a uniform quality-improver across the vocabulary.

The token-class analysis and instruct-sample probe sharpen this caveat.
On the base model, $\lambda$\_MRP=0.6 is broadly positive across every coarse
token class, but structural tokens account for 65.9\% of the net gain.
By $\lambda$\_MRP=2.0, structural share rises to 75.0\%, while content words
fall to 9.4\% and entity-like tokens to 0.9\%. In the single
instruct-style sample, entity-like tokens are net negative ($-$219) even
though overall token accuracy rises by +2.63\%.

The remaining open question is whether this tradeoff persists on
domain-relevant or benchmark-critical token subsets, and whether a
frequency-aware or token-value-aware Fisher weighting can preserve the
head gains without sacrificing medium-frequency content words or
entity-like tokens.

\hypertarget{numerical-precision-implementation-note}{%
\subsection{Numerical Precision Implementation
Note}\label{numerical-precision-implementation-note}}

A separate ablation at 264M scale found that Fisher distance computation
is numerically sensitive to precision: the bf16$\rightarrow$fp32 transition produced
a real regime change in Fisher loss behavior, while fp32$\rightarrow$fp64 was
effectively noise-level. This does not affect the Qwen geometry tables
(all computed in fp32), but it matters for reproducibility: weak Fisher
results under low precision should not be interpreted as evidence
against the method before the Fisher computation path has been verified
in fp32.

\hypertarget{toward-more-precise-margin-refinement}{%
\subsection{Toward More Precise Margin
Refinement}\label{toward-more-precise-margin-refinement}}

The targeted and protected variants (Section 4.13) narrow the next
question. Fisher clearly improves aggregate token accuracy; the open
problem is whether it can be constrained so that gains fall on the
tokens that matter most.

The current evidence suggests not yet. Targeted Fisher remains net
positive, but on the base model it is more head-concentrated than bulk
Fisher $\lambda$\_MRP=0.6, and on instruct-style models the entity-like class is
net negative across all targeted variants. The protection mechanisms are
also insufficient: mean-margin protection does not rescue the
token-value profile, and Elastic Weight Consolidation (EWC; Kirkpatrick
et al., 2017) pilots at 264M scale show no meaningful improvement over
matched no-EWC runs. We therefore view the current targeted/protected
results as evidence that surgical geometric editing is a plausible
research direction, not as evidence that it has been solved.

The practical implication remains important. Domain-tuned models with
ambiguous language or high-value tails may benefit disproportionately
from geometry-aware editing if the objective can be weighted by token
value, frequency, or domain-specific risk. The current pilots expose the
right failure mode and justify further work on token-value-aware Fisher
weighting and better preservation objectives.

\hypertarget{training-time-geometry-shaping-and-scale}{%
\subsection{Training-Time Geometry Shaping and
Scale}\label{training-time-geometry-shaping-and-scale}}

This paper addresses post-hoc polishing, but preliminary results at 264M
and 1.1B scale suggest that training-time geometry shaping --- for
example, margin-oriented objectives during pretraining --- may improve
the manifold substrate on which later Fisher polishing operates by
reducing redundancy and counteracting representational collapse.

Two concrete next steps follow. First, future polishing work should test
whether post-training Fisher retains or destroys improved manifold
structure induced during pretraining. Second, the experiments should be
extended to larger models. Prior work has shown that larger language
models can often be adapted in surprisingly low intrinsic dimension
(Aghajanyan et al., 2021), which suggests more structured slack may be
available for manifold refinement at scale. This is not yet a theorem
about hidden-state manifolds, but it is a credible reason to expect that
larger models may offer more room for geometry-aware refinement than the
4B regime studied here.

\hypertarget{related-work}{%
\section{Related Work}\label{related-work}}

\textbf{Manifold structure in LLMs:} Ansuini et al.~(2019) discovered
the intrinsic dimension hourglass pattern in CNNs. Valeriani et
al.~(2023) extended this to transformers. Mabrok (2026) provided the
first theoretical framework connecting manifold geometry to vocabulary
limitations.

\textbf{Geometric optimization:} Mano (Gu \& Xie, 2026) projects
momentum onto the Oblique manifold tangent space. Both Mano and Muon
(Jordan, 2025) operate on the parameter manifold; our work targets the
representation manifold directly.

\textbf{Confidence calibration:} Guo et al.~(2017) showed modern neural
networks are poorly calibrated. MRP provides a geometric mechanism for
improving calibration: wider margins correspond to more separable
representations, enabling more reliable confidence estimates without
temperature scaling.

\textbf{Speculative decoding:} Leviathan et al.~(2023) showed that draft
model confidence predicts verification success. MRP-widened margins
would improve draft model confidence calibration, potentially increasing
acceptance rates in speculative decoding.

\textbf{Frequency bias and rare-token learning:} Diehl Martinez et
al.~(2024) show that maximum-likelihood pre-training induces frequency
bias and anisotropy, while Chung and Kim (2025) show that
token-frequency imbalance can disproportionately reduce uncertainty on
the most frequent words. Our frequency-stratified Fisher analysis
(Section 4.10) can be read as a post-hoc analogue of this broader
training-time effect.

\textbf{Scale and intrinsic dimension:} Aghajanyan et al.~(2021) showed
that large language models can often be adapted in surprisingly low
intrinsic dimension. That result does not directly measure hidden-state
manifold utilization, but it motivates our suggestion that larger models
may leave more structured slack for geometry-aware refinement.

\textbf{Hallucination detection:} Gao et al.~(2026) identified
hallucination-associated neurons. Our mid-layer geometric ambiguity
finding (Section 4.3) provides a geometric counterpart --- positions
where intermediate-layer ambiguity persists even when the final
prediction is correct.

\hypertarget{conclusion}{%
\section{Conclusion}\label{conclusion}}

We validate Mabrok's latent semantic manifold framework on a
4B-parameter model with R$^2$ = 0.9997, the strongest reported confirmation
of the linear scaling law. We demonstrate that the Voronoi tessellation
of a converged model can be reshaped through targeted post-hoc
intervention --- contrary to the negative results expected from the
MRP-CE redundancy analysis.

The central finding is that Fisher information distance MRP provides a
low-damage mechanism for compressing the expressibility gap in aggregate
metrics across the fully validated range ($\lambda$\_MRP = 0.15 to 0.6). Within
that range, Fisher MRP monotonically improves every geometric metric
while maintaining constant collateral damage (\textasciitilde5,300 R$\rightarrow$W
out of 256K positions) and invariant downstream benchmarks. Additional
runs at $\lambda$\_MRP = 1.0 and 2.0 show continued margin improvement with
near-flat benchmarks, but frequency and token-class audits reveal that
the token-level gains are not uniformly distributed: they are
increasingly concentrated in high-frequency structural tokens, while
content and entity-like contributions shrink as $\lambda$\_MRP rises. A single
fine-tuned instruct-style sample shows the same concentration after
fine-tuning and net-negative entity-like tokens.

The practical implication is therefore narrower than a generic two-phase
recipe. Fisher MRP is best viewed as a post-training geometric polishing
tool whose safest operating point in the current evidence is around
$\lambda$\_MRP=0.6: strong aggregate gains, flat benchmarks, and still-positive
gains across all coarse token classes on the base model. Higher $\lambda$\_MRP
increasingly reallocates utility toward head and structural tokens, and
fine-tuned models may make that tradeoff harsher. The next stage of this
work is not another global-average sweep but token-value-aware
refinement that can preserve the head-token gains without paying for
them with entity-like or medium-frequency content regressions.

The targeted and protected pilots sharpen that recommendation.
Domain-specific or ambiguity-heavy deployments may benefit from more
surgical geometric manipulation, but the current evidence does not yet
show a reliable surgical Fisher recipe. Targeted Fisher remains strongly
head-biased, mean-margin protection is insufficient, and EWC pilots do
not improve the tradeoff. These results are promising as early evidence
that token-value-aware refinement is possible, but do not constitute
finished methods.

\textbf{Limitations and future work.} Several experiments would
strengthen the empirical foundation. (i) \emph{Parameter-scope
ablation:} All experiments use full-model training scope. Because both
Fisher and margin-max use the same scope, the \emph{comparison} between
them (constant vs.~escalating damage) is valid --- the scope is a shared
confounder, not a differential one. However, ablations with frozen
backbone or adapter-only updates would help isolate the minimal
parameter subset responsible for the geometric effects. (ii)
\emph{Uncertainty estimates:} We report single-seed results. While the
dose-response curve provides a form of internal replication ---
monotonic trends across 5--7 $\lambda$ values would not survive high variance
--- formal confidence intervals across seeds would strengthen the
claims. (iii) \emph{Model diversity:} Results are on a single 4B model.
Mabrok (2026) validated the underlying framework across six
architectures; our contribution is the \emph{intervention} study, which
goes deep on one model rather than broad across many. Extension to
additional model families is warranted. (iv) \emph{Calibration metrics:}
Since MRP widens margins, calibration changes (ECE, Brier score) are a
natural test of practical impact beyond flat benchmark means. We have
not yet computed these and expect to include them in a revision. (v)
\emph{Entropy penalty baseline:} We compare Fisher against margin
maximization, but not against an entropy penalty (minimizing H(p) to
sharpen the softmax distribution). An entropy penalty would widen
margins as a side effect of sharpening, without any awareness of
embedding geometry or the Fisher metric. The critical test is whether it
exhibits Fisher's constant-damage property or margin-max's escalating
damage --- our hypothesis is the latter, since entropy sharpening pushes
along the current distribution's gradient without runner-up rotation,
but this remains unverified.

Finally, the polishing result should be read alongside a broader
training hypothesis: training-time geometry shaping may improve the
manifold on which post-hoc Fisher acts. Preliminary results at 264M and
1.1B scale motivate that idea, and prior low-intrinsic-dimension
findings at larger model scales (Aghajanyan et al., 2021) suggest more
room for structured refinement may exist than this paper alone can test.
The most credible long-term path forward is to combine training-time
geometry shaping with post-training token-value-aware polishing rather
than treating bulk Fisher cleanup as the final answer.

\hypertarget{references}{%
\subsection{References}\label{references}}

Aghajanyan, A., Gupta, S., and Zettlemoyer, L. (2021). Intrinsic
Dimensionality Explains the Effectiveness of Language Model Fine-Tuning.
ACL.

Ansuini, A., et al.~(2019). Intrinsic dimension of data representations
in deep neural networks. NeurIPS.

Gao, C., et al.~(2026). H-Neurons: On the Existence, Impact, and Origin
of Hallucination-Associated Neurons in LLMs.

Diehl Martinez, R., Goriely, Z., Caines, A., Buttery, P., and Beinborn,
L. (2024). Mitigating Frequency Bias and Anisotropy in Language Model
Pre-Training with Syntactic Smoothing. EMNLP.

Gu, Y. \& Xie, Z. (2026). Mano: Restriking Manifold Optimization for LLM
Training. arXiv:2601.23000.

Guo, C., et al.~(2017). On Calibration of Modern Neural Networks. ICML.

Kirkpatrick, J., et al.~(2017). Overcoming catastrophic forgetting in
neural networks. PNAS.

Leviathan, Y., et al.~(2023). Fast Inference from Transformers via
Speculative Decoding. ICML.

Mabrok, M. (2026). Latent Semantic Manifolds in Large Language Models.
arXiv:2603.22301.

Merity, S., et al.~(2017). Pointer Sentinel Mixture Models. ICLR.

Chung, W. and Kim, J. (2025). Exploiting Vocabulary Frequency Imbalance
in Language Model Pre-training. NeurIPS.

Qwen Team (2025). Qwen Technical Report. arXiv.

Valeriani, L., et al.~(2023). Geometry of representations in large
transformers. arXiv.

\end{document}